\documentclass[letterpaper, 10 pt, journal, twoside]{ieeetran}

\IEEEoverridecommandlockouts                              

\usepackage{graphics} 
\usepackage{epsfig} 
\usepackage{mathptmx} 
\usepackage{times} 
\usepackage{amsmath} 
\usepackage{amssymb}  
\usepackage{diagbox}
\usepackage{multirow}

\usepackage{tabularx}
\usepackage{booktabs}
\usepackage{bm}
\usepackage{color}
\usepackage[T1]{fontenc}
\DeclareMathOperator*{\argmin}{argmin}

\newcommand{\new}[1]{#1}
\newcommand{\final}[1]{#1}

\title{Robust Body Exposure (RoBE): A Graph-based Dynamics Modeling Approach to Manipulating Blankets over People

}

\author{Kavya Puthuveetil$^{1}$, Sasha Wald$^{1}$, Atharva Pusalkar$^{1}$, Pratyusha Karnati$^{2}$, and Zackory Erickson$^{1}$%
\thanks{Manuscript received: March, 21, 2023; Revised June, 13, 2023; Accepted July, 7, 2023.}
\thanks{This paper was recommended for publication by Editor Markus Vincze upon evaluation of the Associate Editor and Reviewers' comments.
This work was supported by the National Science Foundation Graduate Research Fellowship Program under Grant No. DGE2140739 and by the NVIDIA Academic Hardware Grant.} 
\thanks{$^{1}$ Kavya Puthuveetil, Sasha Wald, Atharva Pusalkar, and Zackory Erickson are with the Robotics Institute, Carnegie Mellon University, Pittsburgh, PA, USA
        {\tt\footnotesize kavya@cmu.edu}}%
\thanks{$^{2}$ Pratyusha Karnati is with Google X, Everyday Robots, CA, USA}%
\thanks{$^{3}$ https://rchi-lab.github.io/robust-body-exposure/}%
\thanks{Digital Object Identifier (DOI): see top of this page.}
}

\begin{document}

\maketitle

\begin{abstract}

Robotic caregivers could potentially improve the quality of life of many who require physical assistance. However, in order to assist individuals who are lying in bed, robots must be capable of dealing with a significant obstacle: the blanket or sheet that will almost always cover the person's body. We propose a method for targeted bedding manipulation over people lying supine in bed where we first learn a model of the cloth's dynamics. Then, we optimize over this model to uncover a given target limb using information about human body shape and pose that only needs to be provided at run-time. We show how this approach enables greater robustness to variation relative to geometric and reinforcement learning baselines via a number of generalization evaluations in simulation and in the real world. We further evaluate our approach in a human study with 12 participants where we demonstrate that a mobile manipulator can adapt to real variation in human body shape, size, pose, and blanket configuration to uncover target body parts without exposing the rest of the body. Source code and supplementary materials are available online$^{3}$.

\end{abstract} 



\section{Introduction}

\IEEEPARstart{F}{or} individuals who have substantial mobility impairments, robots that can provide physical assistance with activities of daily living (ADLs) hold significant promise to improve their independence, agency, and quality of life. Many people require assistance while in bed, often because their disease or injury prevents them from getting out of bed independently. In order to provide care in bed, the first, natural step is to manipulate and uncover blankets that generally will be covering the person's body, which human caregivers are able to do with ease. It is therefore critical for robots to be able to manipulate blankets to uncover parts of the body and enable other assistive tasks to be performed. Robotic caregivers should be able to ensure the privacy and comfort of the person they are assisting by exposing individual limbs or small sections of the body necessary for the completion of the task, just as human caregivers are trained to do~\cite{lynn2011bedbath}, instead of removing the blanket entirely.

Previous work has considered reinforcement learning (RL) and self-supervised learning approaches for targeted bedding manipulation around people~\cite{puthuveetil2022bodies}. However, these model-free methods make several assumptions that limit their practicality and performance. For example, the robot must have a separate model-free bedding manipulation policy for each body part that we aim to uncover. On top of this, any changes to the action space based on the workspace of the robot requires each policy to be retained, which is an expensive process. Furthermore, prior approaches do not include any explicit representation of the cloth so variation in initial cloth state, or in other environmental factors such as human body shape, size, or pose, can degrade performance.

\begin{figure}
    \centering
    \includegraphics[angle=0, width=0.48\textwidth, trim={1.5cm 1.5cm 1.5cm 1.5cm}, clip]{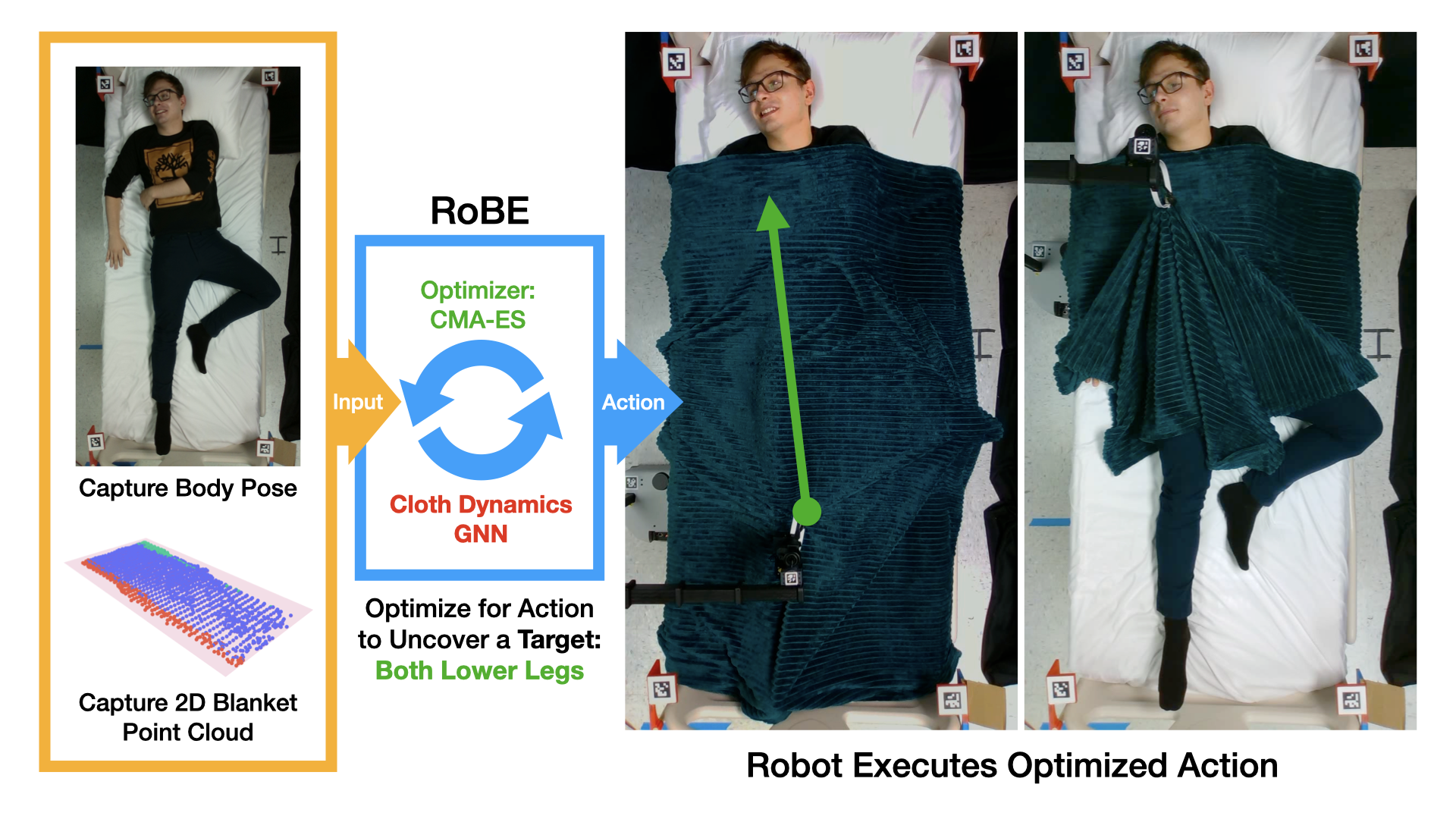}
    \vspace{-0.3cm}
    \caption{\label{fig:title_fig} \new{Using RoBE to expose both lower legs of a participant in the real world using a Stretch RE1 mobile manipulator.}}
    \vspace{-0.6cm}
\end{figure}

In this work, we propose a new method for bedding manipulation, which we call Robust Body Exposure (RoBE), that explicitly reasons about the blanket over the human body, allowing for greater flexibility and improved generalization to variation in the environment. Using data collected in simulation, we first learn a cloth dynamics model that takes in graphs of voxelized point clouds as input. Then, we plan over this model using a representation of the human's pose and body shape in order to uncover a given target, both of which only need to be provided at run-time. 

We show the robustness of RoBE when compared to geometric and reinforcement learning (RL) baselines using a number of in-distribution and generalization evaluations. We further compare our method to these baselines in the real world, deploying all approaches on a real robot to manipulate blankets over a manikin in a hospital bed. \new{Lastly, we conduct a human study with 12 participants to demonstrate RoBE's ability to adapt to both a robot's physical operating constraints and to challenging variations in real human body shape and pose, thus enabling us to uncover a variety of body targets in the real world without additional training on real-world data.} 
For example, Fig.~\ref{fig:title_fig} shows the robot executing a trajectory to uncover both of the lower legs.

Through this work, we make the following contributions:

\final{
\begin{itemize}
\item We introduce a data-efficient and robust approach for targeted bedding manipulation around people that leverages graph-based cloth dynamics modeling.
\item Through a series of evaluations, we quantitatively show that our method is more generalizable than existing reinforcement learning and geometric baselines, and can be easily transferred to real world robots.
\item We present a human study using a mobile manipulator to uncover real people lying in bed. In this study, we demonstrate the effectiveness of our simulation-trained models in manipulating bedding in the real world, subject to real variation of human body shape, body size, and pose, as well as blanket configuration.
\end{itemize}
}

\vspace{-0.2cm}
\section{Related Work}
\subsection{Data-Driven Cloth Manipulation}
Early development of vision-based cloth manipulation strategies focused on leveraging traditional perception-based algorithms to identify cloth features, like corners, wrinkles, and hems, to inform manipulation~\cite{sun2015wrinkles, ramisa2012wrinkles, yamazaki2014hems, yuba2017stateest, qian2020segmentation, seita2019bedmaking}. Since then, a number of data-driven approaches have emerged to improve generalization to different cloth shapes and configurations. These learning-based methods can be broadly defined as either model-free or model-based.

Model-free policies are often trained using imitation learning~\cite{seita2020deep, ganapathi2021learning}, reinforcement learning~\cite{matas2018sim, wu2020learning, lee2021learning, puthuveetil2022bodies}, or learning a value function~\cite{ha2022flingbot} to complete smoothing or folding tasks. While these policies can achieve reasonable task performance on conditions in-distribution, they can struggle to generalize to unseen variation in cloth state, etc.

By contrast, model-based approaches are instead centered on learning a dynamics model of the cloth that can then be used for planning. Some previous work has attempted to learn visual cloth dynamics from RGB-D images~\cite{ebert2018visual, hoque2021visuospatial}.  Contrastive predictive modeling has also been explored for learning a latent representation of cloth dynamics~\cite{yan2021learning}.  Recent advances have implemented particle or mesh-based dynamics models, which, unlike the previous approaches, explicitly reason about the topology of the cloth~\cite{li2018learning, sanchez-gonzalez2020learning, pfaff2020learning, lin2021VCD, huang2022mesh}. While model predictive control (MPC) is used to plan over dynamics models in many of the above approaches, Huang et al. demonstrate how test-time optimization can be used to plan over a mesh-based dynamics model~\cite{huang2022mesh}, a principle that we use in our approach.

\subsection{Cloth Manipulation for Caregiving In and Around Bed}

Despite the great need for robots that can provide assistance to individuals while lying down in bed, previous work in this area is sparse. Kapusta et al. developed an interface that enabled a user with quadriplegia to teleoperate a robot to manipulate a blanket down from the user's knees to his feet~\cite{kapusta2019system}. While making an empty bed is not a physically assistive task, it is still a common home task that people with disabilities may require assistance with. Seita et al. tackle this problem with a self-supervised learning approach that uses depth images to select pick-and-place actions to pull a blanket over a bed~\cite{seita2019bedmaking}.  In our own previous work, we began investigating how blankets could be manipulated over a person lying in bed to uncover target parts of their body via RL and self-supervised learning~\cite{puthuveetil2022bodies}. Seeking to address the shortcomings of our previous approach, including its limited capacity to generalize to a variety of new conditions and expensive training process, we introduce a new formulation of the bedding manipulation task that optimizes over a dynamics model to find bedding manipulation trajectories.

\vspace{-0.2cm}
\section{Model-based Bedding Manipulation}
\label{sec:methods}

\begin{figure*}
    \centering
    \includegraphics[width=\textwidth, trim={0.4cm 3cm 4.8cm 2.5cm}, clip]{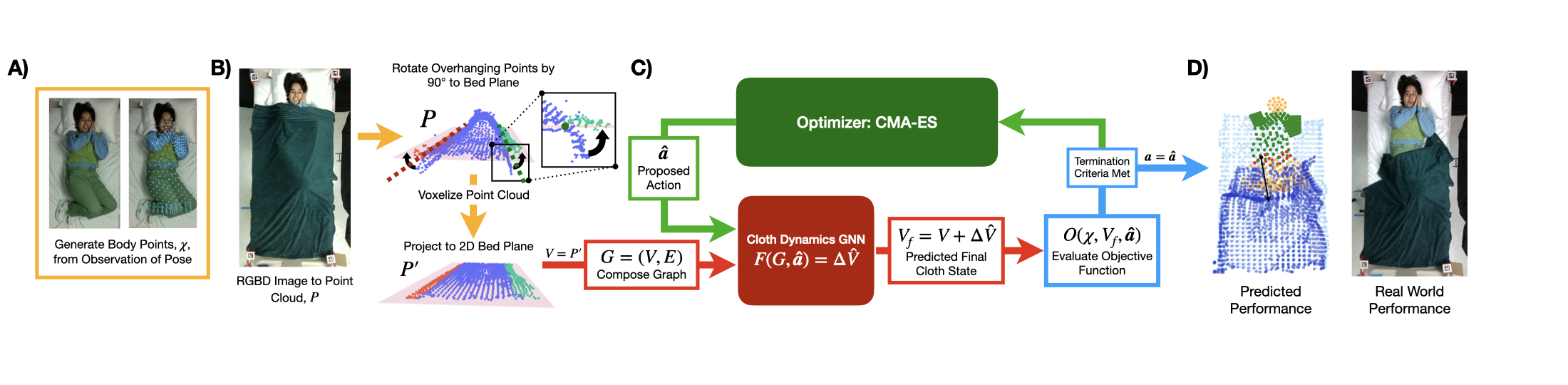}
    \vspace{-0.8cm}

    \caption{\label{fig:pipeline} Overview of our approach: A) We generate a set of body points from an observation of a person's pose. B) We cover them with a blanket, then capture and process an initial cloth point cloud for composition of a graph. C) Given an input graph, we run an optimization loop over the dynamics model to find an action that uncovers a target body part. D) We generate an initial prediction of performance before executing the best action in the real world.}
    
    \vspace{-0.4cm}
\end{figure*}

In this section, we describe our dynamics modeling approach, including our action space, graph representation of cloth, and methodology for collecting data from simulation for model training. We then establish an objective function and discuss how it is used while optimizing over a trained dynamics model to complete the bedding manipulation task. Lastly, we present our set-up for evaluations in the real world. Our approach is summarized in Fig.~\ref{fig:pipeline}.

\vspace{-0.2cm}
\subsection{Dynamics Modeling}

\label{sec:dynamics_modeling}
We model cloth interactions for the bedding manipulation task as 2D, linear pick and place trajectories. These robot actions $\bm{a}$ are 4D vectors containing the 2D position of the grasp (pick) location, $\bm{a}_g = (a_{g,x}, a_{g,y})$, and release (place) location $\bm{a}_r = (a_{r,x}, a_{r,y})$. The action space $A$, where a given action $\bm{a} = (a_{g,x}, a_{g,y}, a_{r,x}, a_{r,y}) \in A$,  is constrained to a bounded planar region corresponding to the top surface of a standard hospital bed such that $a_{g,x}, a_{r,x} \in$ (-0.44~m, 0.44~m) and $a_{g,y}, a_{r,y} \in $ (-1.05~m, 1.05~m).

We represent the state of the cloth as a graph $G = (V, E)$ where the nodes $V = \{v_i\}_ {i=1...N}$ represent particles that compose the cloth and edges $e_{ij} \in E$ represent the connectivity of two nearby particles $v_i$ and $v_j$ on the cloth. For every cloth particle $\bm{p}_i \in P'$ in the voxelized point cloud $P'$ we create a node $v_i \in V$ that encodes the particle's 2D position, $v_i = \bm{p}_i$. A particular edge, $e_{ij} \in E$, exists between given nodes $v_i$ and $v_j$ if the Euclidean distance between nodes is less than some distance threshold, $d$: $ || n_i - n_j||_2 < d$. We also provide a candidate action $\bm{a}$ used to manipulate the cloth as a global vector, which is appended to each node. We encode a 2D projection of a cloth particle's position in the graph instead of its 3D position to ensure depth invariance of models trained using this graph representation. In an evaluation presented on our webpage, we find that both 2D and 3D cloth dynamics models perform similarly when used to manipulate bedding.

Graph neural networks (GNNs) have been well demonstrated for effectively learning cloth dynamics to predict how a robot's actions will perturb a cloth state, particularly using the standard Graph Network-based Simulators (GNS) architecture~\cite{sanchez-gonzalez2020learning, lin2021VCD, wang2022visual}. In this work, we build a GNS-based GNN with 4 processing layers that acts as a dynamics model $F(G, \bm{a}) = \Delta \hat{V}$ that, given a cloth graph $G$ and a candidate action $\bm{a}$, outputs the predicted displacement of all nodes in the graph after execution of the action $\Delta \hat{V}$.

\vspace{-0.2cm}
\subsection{Dynamics Model Training in Simulation}
\label{sec:sim_training}
We train our GNN in simulation with the PyBullet physics engine~\cite{coumans2016pybullet}, which models the physics of planar deformables like cloth as a mass-spring system~\cite{stepian2013physicsanimation}. We interface with PyBullet via Assistive Gym~\cite{erickson2020assistivegym}, a simulation framework for physically assistive robotics. We utilize the same bedding manipulation environment presented in our prior work~\cite{puthuveetil2022bodies}, in which we allow a capsulized human model of fixed body size to settle on a hospital bed in a supine pose with some variation in its joints positions. A blanket, represented by a 1.25m $\times$ 1.7m grid mesh with mass, friction, spring damping, and spring elasticity tuned to match a real-world blanket, is then dropped onto the bed from a fixed position so that most of the human body, with the exception of the head and neck, will be covered. We treat the vertices of the ground truth blanket mesh as points in a point cloud $P$.

\begin{figure}
\centering
 \includegraphics[width=0.12\textwidth, trim={0.6cm 2cm 0.6cm 4cm}, clip]{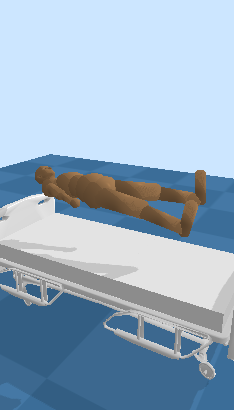} \hspace{-0.2cm}
 \includegraphics[width=0.12\textwidth, trim={0.6cm 2cm 0.6cm 4cm}, clip]{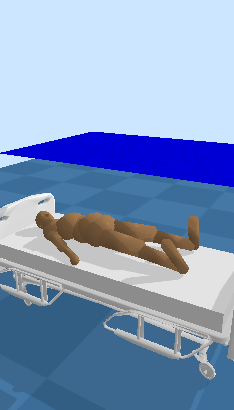} \hspace{-0.2cm}
 \includegraphics[width=0.12\textwidth, trim={0.6cm 2cm 0.6cm 4cm}, clip]{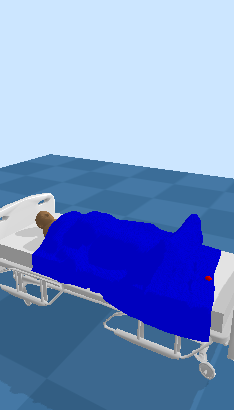} \hspace{-0.2cm}
 \includegraphics[width=0.12\textwidth, trim={0.6cm 2cm 0.6cm 4cm}, clip]{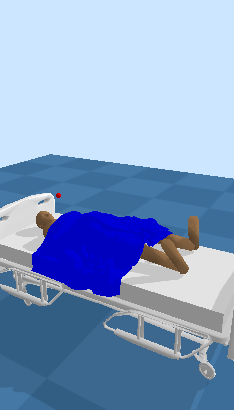} \hspace{-0.2cm}
 \vspace{-0.2cm}

 \caption{\label{fig:simenv} A simulation rollout: 1) We initialize a human model above the bed. 2) After the human model settles on the bed, we initialize a blanket above them. 3) We drop the blanket onto the human model and the end effector picks up the blanket at the grasp location. 4) The blanket is lifted upwards before being moved to the release location.}
 
\vspace{-0.5cm}
\end{figure}

To interact with the cloth given an action $\bm{a}$, we first anchor a spherical manipulator, representing a robot's end effector, to the closest cloth point $p*$ to the grasp point $\bm{a}_g$. We find the cloth point $p^*$ via
$p^* = \argmin_{\bm{p'} \in P \odot (1, 1, 0)} ||\bm{p'} - \bm{a}_g||_2,$
where $\bm{p'}$ represents the 2D position of a cloth point in point cloud $P$ after being projected to a 2D plane, defined by the top surface of the hospital bed, via the Hadamard product, $P \odot (1, 1, 0)$. The manipulator then lifts the cloth off of the human by translating upwards to a fixed height of 40cm to avoid collision with the human body while executing an action. This fixed lift distance corresponds to the maximum lift height of the robot used in our real-world evaluations, described in Section~\ref{sec:real_world_implement}. After lifting the cloth, the manipulator moves linearly between the grasp location, $\bm{a}_g$, to the release location, $\bm{a}_r$, before finally releasing the manipulator-cloth anchor to drop the cloth. An example of a single simulation rollout, from environment setup to trajectory execution is pictured in Fig.~\ref{fig:simenv}.

To collect training data for our dynamics model, we uniformly sample linear trajectories from the action space $A$ to perturb the cloth in a given simulation rollout. If a randomly sampled action contains a grasp point $\bm{a}_g$ that is not over the cloth (i.e., there are no cloth points $p$ within $\lambda = 2.8 cm$ of  $\bm{a}_g$ such that $\exists \bm{p'} \in  P \odot (1, 1, 0):  ||\bm{a}_g - \bm{p'}||_2 < \lambda$), we re-sample actions until the grasp point is valid. We collect a dataset of size 10,000 $(P, \bm{a}, P_f)$ pairs where $P$ is the initial cloth point cloud, $\bm{a}$ is the action, and $P_f$ is the final cloth point cloud after the action is executed. 

For each initial cloth state and action pair in the random interaction dataset, we compose a graph $G = (V, E)$ as described in Section~\ref{sec:dynamics_modeling}, where the nodes $V$ correspond to points in a processed cloth point cloud $P'$. To compute $P'$ from  a given raw cloth point cloud $P$, we first rotate blanket points that hang over the side of the bed up to the bed plane. \new{Rotating overhanging points, as visualized in Fig. ~\ref{fig:pipeline}-B, allows us to more accurately retain the full geometry of the blanket in the graph, which only encodes the 2D position of any particular cloth point.} Our methodology for applying this rotation is described in detail on our webpage. After rotating the overhanging cloth points, we downsample the point cloud using a 5~cm centroid voxel grid filter~\cite{rusu20113d}. Lastly, we project all of the cloth points to the 2D bed plane, yielding the processed point cloud $P'$.




We train a dynamics model $F(G, \bm{a}) = \Delta \hat{V}$, based on the GNS architecture, that given an input graph $G$ that represents a cloth's initial state and a proposed action 
$\bm{a}$, predicts the displacement of the cloth $\Delta\hat{V}$ after execution of a cloth manipulation trajectory. \new{We train our models over 250 epochs on 10,000 graphs---enough training data for performance to saturate---using the Adam optimizer with a mean squared error loss function, batch size of 100, and learning rate of 1e-4. We evaluate our dynamics models when used in the RoBE framework for the bedding manipulation task in Section~\ref{sec:evals}.} 

\vspace{-0.2cm}
\subsection{Determining Covered Body Parts}

\begin{figure}
\centering
\vspace{-0.3cm}
\includegraphics[angle=0, width=0.45\textwidth, trim={2cm 12cm 2cm 12cm}, clip]{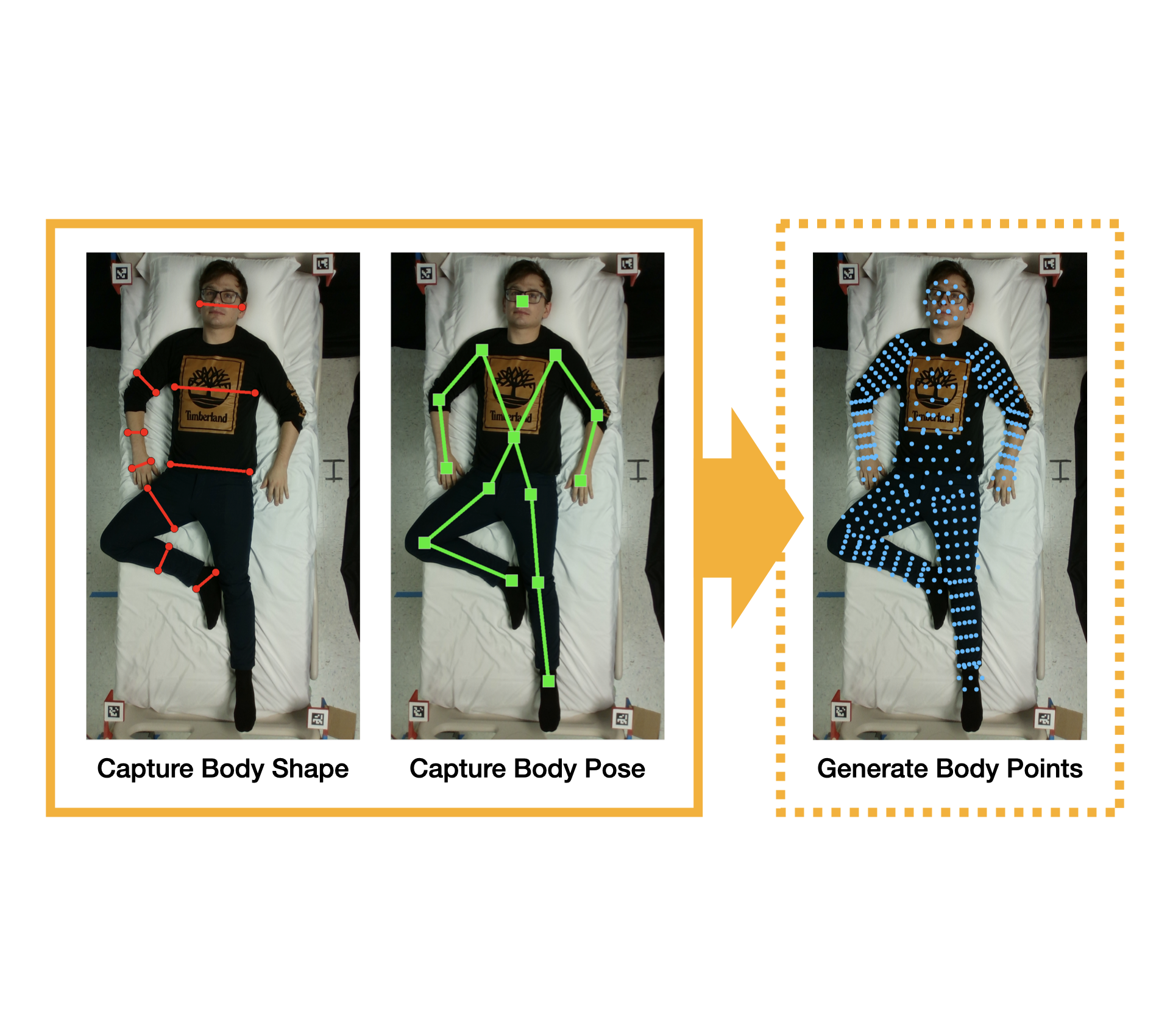}
\vspace{-0.3cm}
\caption{\label{fig:body_info_pose_est} Capturing human body shape parameters and pose in the real world: 1) Measuring each of the nine body segments, 2) Adjusting the estimated position of the 14 joints in observation $\bm{s}$ to better match the ground truth pose, 
3) visualization of all the virtual points generated along the body.}
\vspace{-0.5cm}
\end{figure}

To assess whether areas on the body are covered or uncovered, we represent the human body using a set of 384 discrete, uniformly spaced, points $\chi$. To generate this set of points, we capture a 28D observation of the human's pose, $\bm{s}$, defined by the 2D positions of 14 different joints. We also use a set of nine body shape measurements that includes the radius of several body segments. The 14 joint positions and the nine body shape measurements we use are pictured in Fig.~\ref{fig:body_info_pose_est}. We uniformly distribute points across the area of each body segment, as calculated using their length and radius/width, to build our representation of the human.



From a set of human body points $\chi$, we can determine whether a given body point $x \in \chi$ is covered or uncovered by the final state of the cloth $V_f = V + \Delta V$ via the function:
\begin{equation}
\label{eqn:C}
C(x, V_f) = \begin{cases} 0 & \text{if }\exists v_f \in V_f :  || x - v_f||_2 < \lambda\\1 & \text{otherwise}\\\end{cases}
\end{equation}
which states that a body point $x \in \chi$ is considered covered $(C(x, V_f) = 0)$ if there are any cloth points within $\lambda=2.8$cm of it, otherwise it is considered uncovered $(C(x, V_f) = 1)$. The threshold $\lambda$ was tuned manually in previous work~\cite{puthuveetil2022bodies}.

\vspace{-0.2cm}
\subsection{Optimizing over Dynamics Models}
\label{sec:optimizing}
Following training, we use our dynamics models to predict how the state of a cloth will change when manipulated using a proposed action. Compared to physics simulation, forward passes of our GNNs can be run within a fraction of a second. This enables them to be used in optimizing for actions that generate favorable cloth states in close to real time. Favorable cloth states uncover a target body part without exposing the rest of the body. We define ten discrete targets: \textit{the right arm, right lower legs, right legs, left arm, left lower legs, left legs, both lower legs, upper body, lower body, and entire body}. The set of all body points $\chi$ can be further divided by whether they are target body points $\chi_t$, non-target body points $\chi_n$, or points on the head $\chi_h$, such that $\chi_t, \chi_n, \chi_h \subseteq \chi$

In order to optimize for actions that uncover the body in a targeted way, we define an objective function:
$$O(\chi, V_f, \bm{a}) = \alpha O_t(\chi_t, V_f) + \phi O_n (\chi, V_f)+ O_h(\chi_h, V_f) + O_d(\bm{a})$$

These terms represent, in order, a reward for uncovering target points and penalties for uncovering non-target points, covering the head, or choosing excessively long actions. $\alpha = 100$ and $\phi = 100$ are scalar weights that can be used to modify the reward for uncovering target points and penalty for uncovering non-target points, respectively.


The reward for uncovering target body points is given by: $O_t(\chi_t, V_f) = -\alpha \:(\rho_t/|\chi_t|)$ where $\rho_{t} = \sum_{x\in \chi_{t}} C(x, V_f)$ counts target points uncovered and $|\chi_t|$ is the cardinality of set $\chi_t$.

The penalty for uncovering non-target body points is:
\begin{equation}
\label{eqn:nt_penalty}
O_n(\chi, V_f) = \sum_{x\in \chi_{n}} C(x, V_f) * \Gamma(x, \chi_t) * \Xi(\chi, \chi_t)
\end{equation}
where $\Gamma(x, \chi_t) = \argmin_{x^* \in \chi_t} ||x^* - x||_2$ is a scaling function that adjusts the penalty weight of a given non-target point $x \in \chi_n$ based on its distance to the closest target point $x^* \in \chi_t$. This term operates on the intuition that uncovering non-target points close to the target limb may be inevitable in some poses. We reason that uncovering such points should not incur as much cost as uncovering those that are farther away from the target (i.e. are theoretically easier to avoid uncovering). We also account for the increased difficulty of uncovering smaller targets with
the scaling term $ \Xi(\chi, \chi_t) = \frac{|\chi_t|}{0.05 * (|\chi_n| + |\chi_h|)}$, which further adjusts the penalty for uncovering a non-target body point based on the relative size of the target, defined as the ratio between the number of target points and the number of non-target and head points.

We also include a heavy penalty for covering parts of the head, $O_h(\chi_h, V_f) = 2\alpha\: (\rho_h/|\chi_h|)$ where $\rho_{h} = |\chi_h| -  \sum_{p\in \chi_{h}} C(p, V_f) $, since doing so is generally against human preferences. 
Lastly, $O_d(\bm{a}) = 150$ is a penalty applied when the Euclidean distance between grasp and release points, $||\bm{a}_r - \bm{a}_g||_2$, is greater than 1.5m. Otherwise, $O_d(\bm{a}) = 0$.

With a trained dynamics model and an objective function, we can introduce an optimizer, in our case covariance matrix adaptation evolution strategy (CMA-ES)~\cite{hansen2003reducing}, to identify optimal bedding manipulation trajectories to uncover people in bed in real time. Given an initial processed cloth point cloud $P'$, on each iteration of CMA-ES, a graph $G$ is constructed with a new sampled action, $\hat{\bm{a}}$, and then fed into the dynamics model to predict the displacement of the cloth $\Delta \hat{V}$. We then compute the final cloth state $V_f = V + \Delta \hat{V}$, construct the set of all points along the body $\chi$ from an observation of human pose, define $\chi_t, \chi_n, \chi_h \subseteq \chi$ for a specific target limb, and finally evaluate objective function $O(\chi, V_f, \hat{\bm{a}})$ before re-sampling a new set of actions based on the cost value. We allow CMA-ES to optimize over the action space $A$ for 38 iterations using a population size of 8 (304 total function evaluations), or until $O(\chi, V_f, \bm{a}) \geq 95$. Optimization hyperparameters are further described on our webpage. 

Compared to other bedding manipulation methods, our strategy of optimizing over a dynamics model allows us to uncover new targets, modify objectives, or change the importance of sub-objectives, all in real time without needing to retrain. For example, we can, without retraining, restrict the optimizer search space to a given robot's workspace, which we demonstrate in Section~\ref{sec:real_world_implement}, or modify weights $\alpha$ and $\phi$ in $O(\chi, V_f, \bm{a})$ to prioritize uncovering the entirety of a target with less concern for uncovering non-target points.





\subsection{Real World Implementation}
\label{sec:real_world_implement}
We implemented RoBE using a Stretch RE1 mobile manipulator~\cite{kemp2021designofstretch} to demonstrate its efficacy in the real world.

In order to capture a complete point cloud of the blanket, including points hanging over the sides of the bed, we use three cameras affixed above, and on the right and left sides of, the bed. After capturing point clouds from each camera and applying color segmentation and position-based cropping on the point clouds to isolate the points on the blanket, we perform Iterative Closest Point (ICP) registration over 5e6 iterations to reconstruct a single point cloud of the blanket.

\new{Before manually covering the human with the blanket, we capture an RGB image using a camera mounted above the center of the bed.} We feed this image into BlazePose, as implemented in Google's MediaPipe Pose~\cite{bazarevsky2020blazepose}, to capture an estimate of the human's joint positions. \new{Although we currently rely on pose estimation methods that require an RGB-based, unobstructed observation of the human body, future work could instead use occlusion-invariant methods that use images from a bed pressure mat~\cite{clever2018pressurepose} or from an depth camera~\cite{clever2021depthpose}. Such methods may also help alleviate concerns associated with the use of RGB cameras in private areas.} Using a custom GUI, we manually adjust pose estimates from BlazePose for better alignment with the body and also take body shape measurements, thus allowing us to generate the set of body points $\chi$. Fig.~\ref{fig:body_info_pose_est} depicts this process. 

\begin{figure}
\centering
\includegraphics[angle=0, width=0.45\textwidth, trim={1cm 18cm 1cm 10cm}, clip]{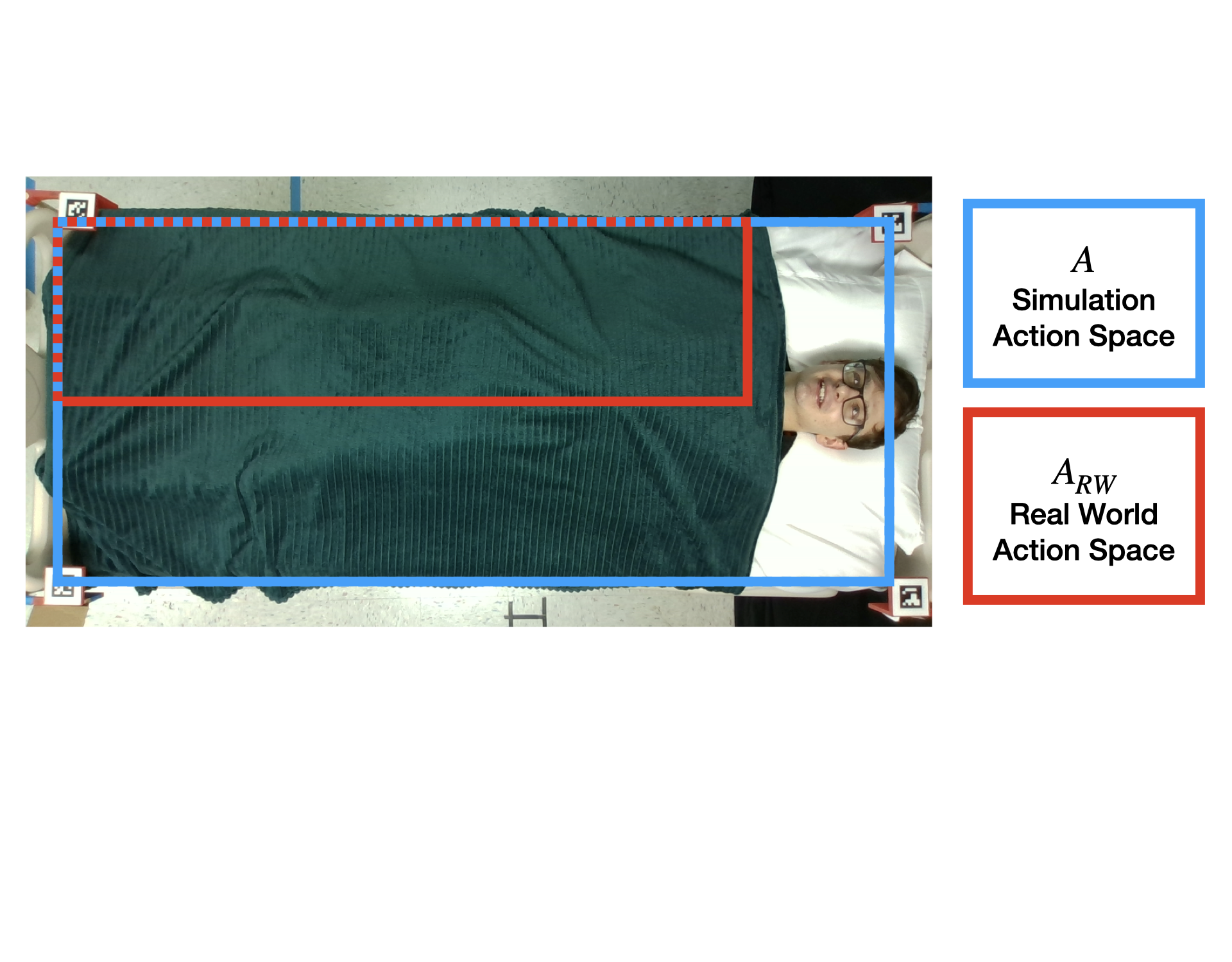}
\vspace{-0.3cm}
\caption{ \label{fig:action_spaces} Comparison of the action space used in simulation, $A$, and the action space used in the real world implementation, $A_{RW}$.}
\vspace{-0.5cm}
\end{figure}

Using fiducial tags fixed to each of the corners of the bed, we find the center of the real-world hospital bed, allowing us to transform the point cloud of the blanket and the human pose estimate to a planar global coordinate system comparable to that used in simulation. \new{We apply the same point cloud preprocessing steps as in simulation, including rotation along bed edges, voxelization, and 2D projection. This allows us to construct an input graph to the dynamics model and optimize the objective function, as described in Section~\ref{sec:optimizing}, to find an action that uncovers a specific limb without fine-tuning.} The Stretch robot's maximum arm extension (52~cm) does not allow it to reach across a hospital bed (88~cm wide), however, our approach can easily accommodate the robot's physical limitations by simply restricting the search space for CMA-ES to the robot's workspace, which does not require any retraining. We also restrict the search space to prevent the robot from grasping near participants' head or neck, which could cause discomfort. The resulting action space $A_{RW}$ used in the real-world experiment is defined such that $a_{g,x},a_{r,x} \in$ (0~m, 0.44~m), $a_{g,y} \in $ (-0.525~m, 1.05~m), and $a_{r,y} \in $ (-1.05~m, 1.05~m), as shown in Fig.~\ref{fig:action_spaces}. To account for the robot's limited reach, we only attempt to uncover seven of the ten targets enumerated in Section~\ref{sec:optimizing} in the real world, excluding the left arm, lower leg, and leg.


\begin{figure}
    \centering
    \includegraphics[width=0.45\textwidth, trim={9.5cm 1cm 9.5cm 2cm}, clip]{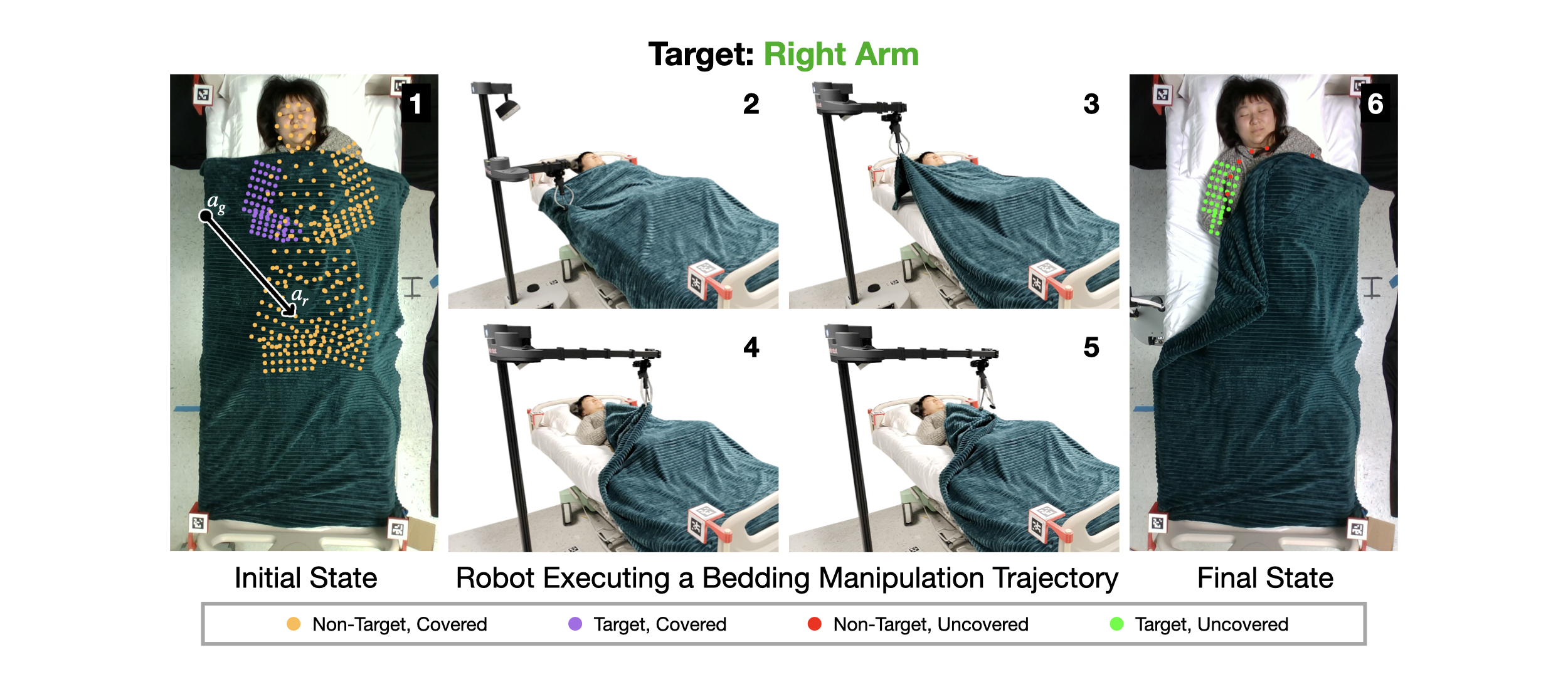}
    \vspace{-0.3cm}
    \caption{\label{fig:robot_execution} Executing RoBE to uncover a participant's right arm: 1) The initial pose of the participant and action to be executed. 2) The robot grasps the cloth at the grasp locations, 3) lifts the cloth, 4) linearly moves the cloth to the release location, 5) and drops the cloth. 6) The final state of the cloth and the target and non-target points that have been uncovered.}

    \vspace{-0.4cm}
\end{figure}

We provide computed actions to the robot for execution. At the beginning of a given trial, the robot uses its pan-tilt camera to observe the fiducial markers on the corners of the bed and localize itself to the real-world global coordinate system that the grasp and release points are defined with respect to. The robot moves its end effector to the grasp location with its gripper closed, then lowers the gripper until it contacts the blanket, as detected by effort sensing in the robot's actuators, to complete the grasp. It lifts up by 3cm to open its gripper, lowers to the cloth to close its gripper on the blanket, and then raises its end effector to 40cm above the bed before moving linearly to the release location to drop the blanket. Lastly, we capture an image from the above-bed camera to use in evaluation of the robot's performance. This sequence of trial events is shown in Fig.~\ref{fig:robot_execution}. \new{Completion of a real-world trial, from capturing pose to executing an action, takes 4-5 min, of which the robot motion takes $\sim$1 min.}

\section{Evaluations}
\label{sec:evals}
\subsection{Evaluation Metrics}
We perform experiments both in simulation and in the real world to assess the efficacy of our method. We define and compute the following metrics to assess task performance:

\begin{itemize}
\item $\text{True positive }(TP) = \rho_{t}$
\item $\text{False negative }(FN) = |\chi_{t}| - \rho_{t}$
\item $\text{False positive }(FP) = \gamma(\rho_n)$
\item $\text{F-Score } (F_1) = \frac{TP}{TP + 0.5(FP+FN)}$
\end{itemize}

TP is the number of target body points that are correctly uncovered while FN is the number of target points that remained covered. FP is a weighted representation of the number of non-target points that are uncovered, $\gamma(\rho_n)$:

$$ \gamma(\rho_n) = \sum_{i=1}^{\rho_n} \begin{cases} w*i & \text{if } w * i \leq 1\\
1 & \text{otherwise}\\\end{cases}$$
where $\rho_{n} = \sum_{p\in \chi_{n}} C(p, V_f)$ is the total number of non-target points uncovered and $w = \frac{|\chi_t|}{|\chi|}$ is a weighting factor representing the size of a target limb relative to the size of the entire body. 
Similar to Equation~\ref{eqn:nt_penalty}, $\gamma(\rho_n)$ operates on the intuition that, when evaluating a bedding manipulation outcome, people are willing to accept that some non-target body points may inevitably be uncovered, like when uncovering regions of the body that are spatially close to other regions (e.g. uncovering an arm that is resting on the stomach). 


Finally, $F_1$ is a weighted measure of the robot's accuracy at performing bedding manipulation, where $F_1 \in [0, 1]$.

\subsection{Baselines}
We compare RoBE to two baselines, including a geometric approach and a reinforcement learning method:

\subsubsection{Geometric Baseline}
This approach is founded on two assumptions that: 1) there exists a geometric relationship between the cloth and human pose that can be leveraged to uncover a blanket from part of the body and, 2) the cloth behaves like paper---pick and place actions can produce straight-line or accordion folds. To compute actions that ``fold'' the cloth over the body to uncover a target limb, we apply three strategies depending on the target to be uncovered; \textit{Arms:} grasps the edge of the cloth, pulls towards the midline of the body on a path normal to the length of the arm;  \textit{Lower Body Targets:} grasps along the midpoint of the target, pulls towards the head on a path parallel to the limb; \textit{Upper Body and Entire Body:} grasps between the shoulders, pulls toward the feet on a path that bisects the body. We provide formal definitions of how each strategy computes these pick and place trajectories on the project webpage.

\subsubsection{Reinforcement Learning}
For this baseline, we use proximal policy optimization (PPO), a model-free deep RL method, to train policies to uncover a single target limb as demonstrated in prior work~\cite{puthuveetil2022bodies}. We train a single policy for each of the ten target limbs, using $-O(\chi, V_f, a)$ as a reward function and the 28D representation of human pose $\bm{s}$ as the observation, similar to prior work~\cite{puthuveetil2022bodies}. Running PPO in the real world required the expensive process of training a second set of policies for each target limb, this time using the action 
space $A_{RW}$.

We evaluate all methods both in simulation, described in Section~\ref{sec:sim_evals}, and in the real world with a medical manikin, described in Section~\ref{sec:manikin_study}.

\subsection{Evaluating Generalization in Simulation}
\label{sec:sim_evals}

\new{We conduct all simulation evaluations over 500 rollouts where both the environment's state and the target limb are randomized over the following generalization conditions:}

\subsubsection{Training (Training Dist.)} The model training conditions (Section~\ref{sec:sim_training}), including some small human pose variation with fixed body size and initial blanket configuration.

\subsubsection{Human Body Shape and Size Variation (Body Dist.)} For this condition, we introduce large variation to the human body shape and size by fitting the capsulized human model to a SMPL-X body mesh~\cite{pavlakos2019expressive}, defined using 10 uniformly sampled body shape parameters $\beta \in U(0, 4)$, before dropping the model onto the hospital bed as described in~\cite{puthuveetil2022bodies}.

\subsubsection{High Human Pose Variation (Pose Dist.)} We apply additional pose variation beyond that used in the training distribution by adding a $\delta$ to the joint angles of both of the capsulized model's elbows and knees. For each of these four joints, this $\delta$ value is uniformly sampled from the range $(\theta_{min} + 1~\text{rad}, \theta_{max} - 1~\text{rad})$, where $\theta_{min}$ and $\theta_{max}$ are the limits of a given joint's range of movement~\cite{erickson2020assistivegym}.

\subsubsection{Initial Blanket Variation (Blanket Dist.)} We modify the initial blanket configuration by introducing variation uniformly sampled between $\pm$2cm to its $x$ position, $[-25, 5]$ cm to its $y$ position, and $\pm$45 degrees to its yaw $\theta_z$ orientation before dropping the blanket onto the human.

\subsubsection{Combination (Combo. Dist.)} We use all variations from the Body, Pose, and Blanket Dists., resulting in challenging environmental states relative to the Training Dist. 


    

\begin{table}
\centering
\caption{\label{table:sim_and_manikin_eval} Generalization performance ($F_1$) in simulation and in a real-world manikin study}
\begin{tabular}{cccc} \toprule
    \multirow{2.5}{*}{Evaluation Conditions} & 
    \multicolumn{3}{c}{Approaches} \\ \cmidrule{2-4}
    & RoBE&PPO&Geometric \\ \midrule\midrule

    Training Dist.   & $\bm{0.77\pm0.23}$ & 0.25$\pm$0.40 & 0.41$\pm$0.23\\
    Body Dist.            & $\bm{0.72\pm0.25}$ & 0.31$\pm$0.40 & 0.43$\pm$0.23\\
    Pose Dist.            & $\bm{0.67\pm0.29}$ & 0.26$\pm$0.38 & 0.39$\pm$0.26\\
    Blanket Dist.            & $\bm{0.72\pm0.25}$ & 0.27$\pm$0.39 & 0.39$\pm$0.25 \\
    Combo. Dist.            & $\bm{0.65\pm0.30}$ & 0.33$\pm$0.38 & 0.41$\pm$0.31\\
    \midrule
    Manikin Study & $\bm{0.81\pm0.12}$ & 0.23$\pm$0.32 & 0.63$\pm$0.16\\
	\bottomrule
\end{tabular}
\end{table}

Performance of our approach and both baselines on each generalization condition is summarized in Table~\ref{table:sim_and_manikin_eval}. We find that across all generalizations, our method achieves the highest F-scores, followed by the geometric approach, and PPO respectively. The poor performance of PPO policies can be attributed to them often selecting trajectories that induce little to no displacement of the cloth (e.g. not grasping the cloth at all, picking up and releasing the cloth in effectively the same place). The geometric approach does slightly better, but its predefined strategies limit its ability to generalize to a larger variety of poses and blanket configurations. 

\begin{figure}
\centering
\includegraphics[angle=0, width=0.43\textwidth, trim={8cm 1cm 8cm 3cm}, clip]{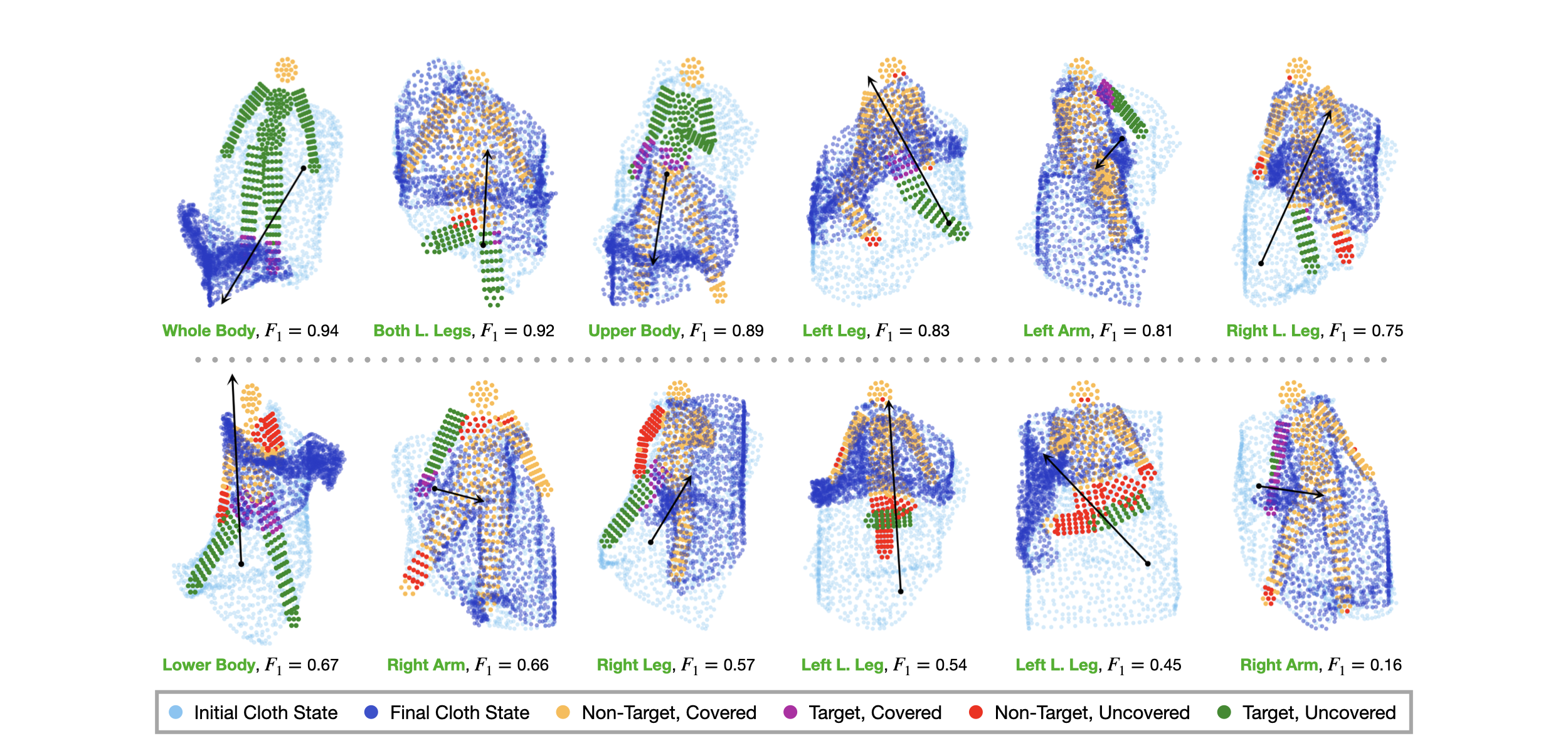}
\caption{ \label{fig:combo_var_imgs} Representative rollouts of our approach when evaluated on the combination distribution.}
\vspace{-0.5cm}
\end{figure}

By contrast, our approach leverages a learned model that is close to the cloth's true dynamics when optimizing for actions. RoBE's F-scores vary from 0.65-0.72 on generalization conditions compared to 0.77 on the training distribution. Fig.~\ref{fig:combo_var_imgs} shows representative samples of how RoBE performs on the combination distribution and provides context to how F-score varies depending on the manipulation outcome. \new{We observe decreased dynamics model accuracy when the cloth's dynamics are more complex (Pose, Combo. Dists.), thus degrading task performance. Future work may consider how reasoning about the underlying connectivity of the blanket can improve performance in these cases~\cite{lin2021VCD, huang2022mesh}.} As shown in Fig.~\ref{fig:combo_var_imgs}, we also note that some target limbs are near impossible to uncover in isolation due to the human body shape, pose, and initial blanket state.

\vspace{-0.2cm}
\subsection{Manikin Study}
\label{sec:manikin_study}
We compare our approach to baselines in the real world by deploying the robot to uncover target limbs of a medical manikin lying in the hospital bed. In the study, we evaluate all three methods for uncovering each of the seven target limbs on a set of three poses (3 methods $\times$ 3 poses $\times$ 7 targets = 63 total trials). Table \ref{table:sim_and_manikin_eval} summarizes the performance of each approach in the real world over the manikin. These results are generally consistent with what we observed in simulation, with RoBE ($F_1 = 0.81$) outperforming the geometric approach by 0.18 and PPO by 0.58. 

\vspace{-0.2cm}
\subsection{Human Study}
\label{sec:real_world_eval}
\begin{figure}
\centering
\includegraphics[angle=0, width=0.45\textwidth, trim={12cm 2cm 12cm 5cm}, clip]{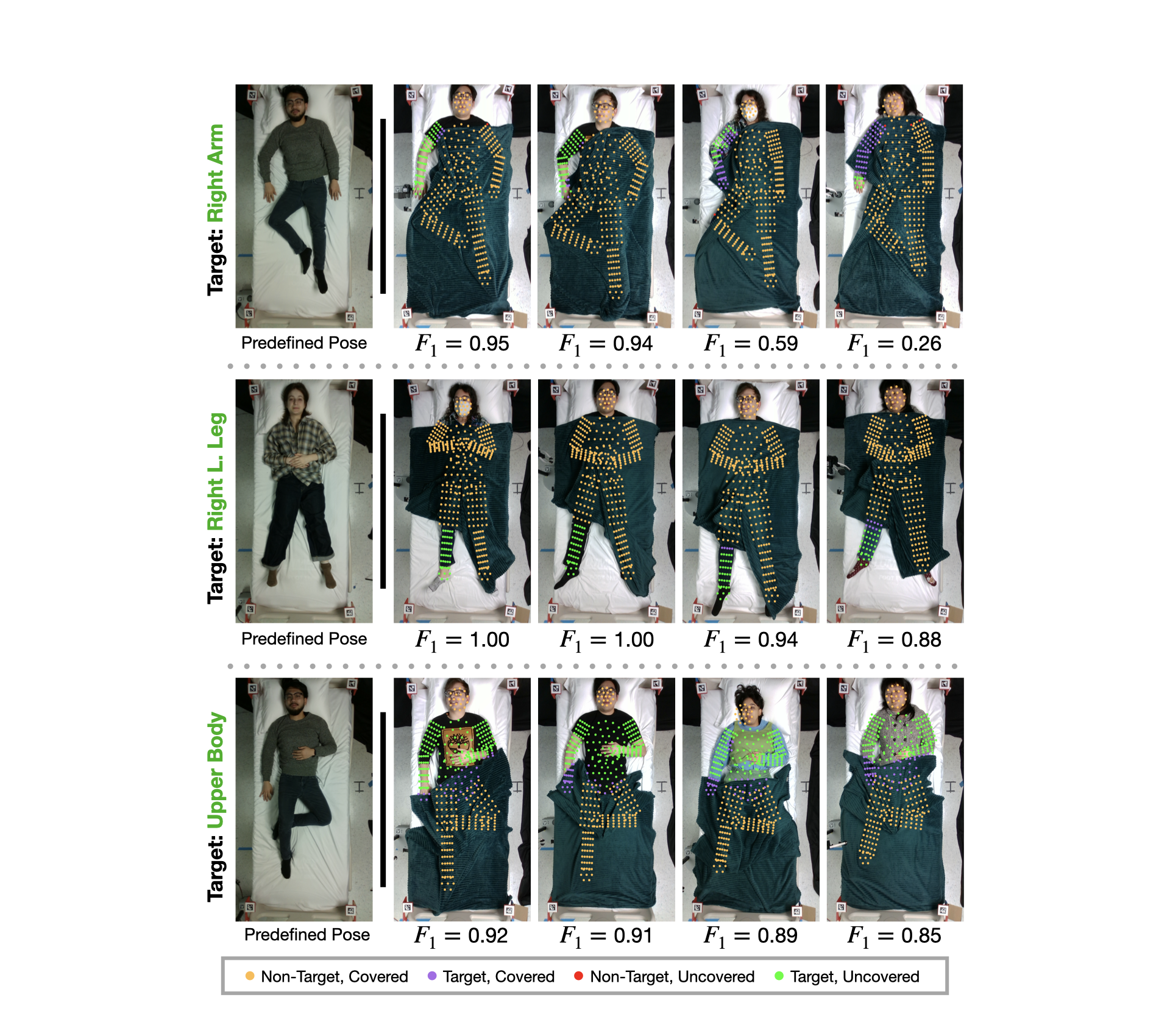}
\vspace{-0.3cm}
\caption{ \label{fig:predef_poses} Representative performance from the first phase of the human study where both the target limb and participant's pose are predefined. Images of the predefined poses shown to participants are in the leftmost column. Results are ordered in decreasing F-score.}
\vspace{-0.5cm}
\end{figure}

To evaluate the performance of our method in the real world, we run a human study (Carnegie Mellon University IRB Approval under 2022.00000351) with 12 participants (six female) with informed consent. For each participant, we demonstrate ten bedding manipulation trials (120 trials in total across participants). The ten trials are executed in two phases. The first phase consists of three trials where the participant is asked to assume a predefined pose, shown in the leftmost column of Fig.~\ref{fig:predef_poses}, and the robot attempts to uncover a predefined target body segment. By holding the pose and target constant, we examine the performance of our method given variation in body shape and size alone. The remaining seven trials are a part of Phase 2, in which we ask participants to assume any supine pose of their choosing and we randomly select a target body part to uncover. This second phase allows us to evaluate the robustness of our approach to a large set of unscripted resting poses. \new{Since our approach operates under the assumption that the human's pose is static, we ask participants to stay still during a given trial.}

\begin{figure}
\centering
\includegraphics[angle=0, width=0.46\textwidth, trim={5cm 6cm 6cm 11.5cm}, clip]{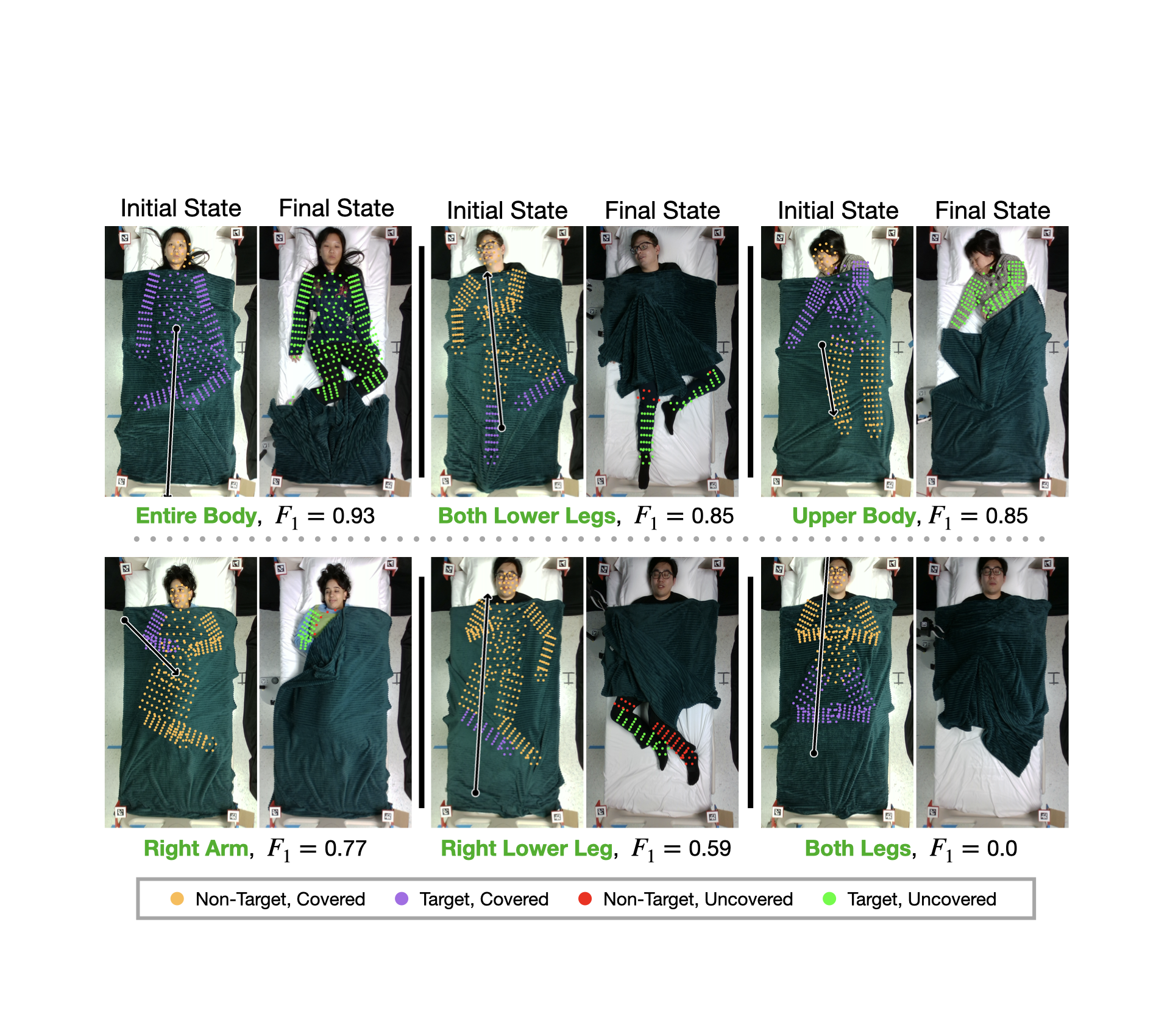}
\vspace{-0.3cm}
\caption{Example images from the second phase of the study where participants chose their own supine poses and the target limb was randomized. Results are ordered in decreasing F-score.\label{fig:human_study_random_pose} }
\vspace{-0.5cm}
\end{figure}

The mean F-score achieved by our method across target limbs in each phase of the human study is summarized in Table~\ref{table:human_study_eval}. For comparison, we provide the detailed generalization results of our approach in simulation using the Combo. Dist. A representative selection of trials from Phase 1 and Phase 2 are visualized in Fig.~\ref{fig:predef_poses} and Fig.~\ref{fig:human_study_random_pose}, respectively.

\begin{table}
\centering
\caption{\label{table:human_study_eval} Performance ($F_1$) Per Limb in the Real World and Simulation}
\begin{tabular}{ccccc} \toprule
    \multirow{2.5}{*}{Target Body Part} & 
    \multicolumn{2}{c}{Human Study} & & Simulation \\ 
    \cmidrule{2-3} \cmidrule{5-5}
    
    & Predefined & Randomized && Combo Dist. \\ \midrule\midrule
    
    Right Arm       &   0.77$\pm$0.29     &     0.70$\pm$0.26    &&   0.50$\pm$0.32\\
    Right Lower Leg &   0.95$\pm$0.05     &     0.51$\pm$0.33    &&   0.57$\pm$0.29\\
    Upper Body      &   0.88$\pm$0.05     &     0.89$\pm$0.12    &&   0.87$\pm$0.12\\
    Right Leg       &           ---       &     0.51$\pm$0.24    &&   0.52$\pm$0.22 \\
    Both Lower Legs &           ---       &     0.67$\pm$0.27    &&   0.77$\pm$0.18 \\
    Lower Body      &           ---       &     0.61$\pm$0.28    &&   0.84$\pm$0.10 \\
    Entire Body     &           ---       &     0.87$\pm$0.05    &&   0.96$\pm$0.09 \\
    \midrule
    Overall (Mean)  &         ---         &     0.68$\pm$0.27    &&   0.65$\pm$0.30\\
	\bottomrule
\end{tabular}
\vspace{-0.5cm}
\end{table}

In the Phase 1 trials, where both pose and target limbs were predefined, we observe strong performance when uncovering the right lower leg and upper body, with high mean F-scores ($ \geq 0.88$) and low standard deviations ($0.05$), indicating that we can reliably uncover these targets despite variation in human body shape and size. We see lower performance ($ F_1 = 0.77$) in uncovering the right arm. The increased standard deviation relative to the other two targets suggests that the right arm is a more challenging target to uncover for some body sizes. Yet, overall we observe that the generalization of our approach across real variation in body shape and size is promising.

In Phase 2, we evaluated our approach on a variety of challenging scenarios where participants selected their pose and we randomized the target. We observed high performance consistent with simulation for most body targets.


Of the failure cases we observed, we found that many were less due to errors in our model and more attributable to the robot failing to complete a given trajectory. Topographical features created by the body that are normal to the bed's surface, such as upturned feet or knees bent upwards, could generate high frictional forces as the robot attempts to drag the blanket over them. For longer trajectories, these forces can become so high that they can keep the robot from advancing forward or can pull the blanket out of the robot's gripper. A failure case demonstrating an example of this phenomenon is shown the bottom right in Fig.~\ref{fig:human_study_random_pose}.

Although using a robot that can pull at higher forces could potentially resolve this problem, there may be safety concerns with doing so. Instead, future work may consider incorporating feedback from estimated force and contact maps, as has been previously demonstrated for robot-assisted dressing~\cite{erickson2017does, erickson2018deep, wang2022visual}, to modify or inform bedding manipulation trajectories. Aside from some limitations with our mobile manipulator, the results achieved in the human study with our current approach remain promising towards achieving robust bedding manipulation in the real world.

\section{Conclusion}
In this work, we leverage gradient-free optimization over simulation-trained graph-based cloth dynamics models to inform targeted bedding manipulation over people. One of several benefits of this strategy is the capability to change the objective function and goals in real time without retraining. Compared to geometric and reinforcement learning baselines, our approach is robust and generalizable, as demonstrated by a battery of evaluations in simulation and in the real world with a medical manikin. In a human study, we also demonstrate RoBE's efficacy when performing targeted bedding manipulation over real people. We observed that our approach had promising performance uncovering targeted body regions across different human body sizes, variations in human pose, and different blanket configurations.

\addtolength{\textheight}{-12cm}   









\bibliographystyle{IEEEtran}
\bibliography{root}

\end{document}